\documentclass[letterpaper,10pt]{article}

\usepackage[letterpaper,margin=0.75in]{geometry}
\usepackage{graphicx} 
\usepackage{amssymb}
\usepackage{amsmath}  
\usepackage{fancyhdr} 
\usepackage[compact]{titlesec} 
\usepackage[numbers,sort&compress]{natbib} 
\usepackage{geometry}
\usepackage[inline]{enumitem}

\usepackage{tikz}						
\usepackage{circuitikz}
\usetikzlibrary{trees}
\usetikzlibrary{calc}				
\usetikzlibrary{plotmarks}
\usepgflibrary{arrows}
\usetikzlibrary{positioning}
\usetikzlibrary{shapes.arrows}
\usetikzlibrary{positioning,calc,shapes,arrows,shapes.multipart}
\tikzset{
	papDecision/.style = {
		diamond,
		draw, 
		text width = 18 mm, 
		align = center, 
		text badly centered,
		inner sep = 1 pt,
		font=\footnotesize,
		minimum width = 25mm,
		minimum height = 7mm,
	},
	papStart/.style = {
		rectangle,
		draw, 
		align = center, 
		text width = 3.3cm, 
		text badly centered,
		inner sep = 4 pt,
		rounded corners=10pt,
		font=\footnotesize,
		minimum width = 30mm,
		minimum height = 7mm,
	},
	papEnd/.style = {
		rectangle,
		draw, 
		align = center, 
		text width = 3.3cm, 
		text badly centered,
		inner sep = 4 pt,
		rounded corners=10pt,
		font=\footnotesize,
		minimum width = 30mm,
		minimum height = 7mm,
	},
	papData/.style = {
		trapezium,
		draw, 
		align = center, 
		text width = 20 mm, 
		text badly centered,
		inner sep = 4 pt,
		trapezium left angle=70,
		trapezium right angle=110,
		font=\footnotesize,
		minimum width = 30mm,
		minimum height = 7mm,
	},
	papPredProc/.style = {
		draw,
		rectangle split,
		rectangle split horizontal,
		rectangle split parts = 3,
		rectangle split empty part width=-8pt,
		align = center, 
		text badly centered,
		font=\footnotesize,
		minimum width = 30mm,
		minimum height = 7mm,
	},
	papProcess/.style = {
		rectangle,
		draw,
		align = center, 
		text width = 3.3cm, 
		text badly centered,
		font=\footnotesize,
		minimum width = 30mm,
		minimum height = 7mm,
	},
	papLine/.style = {
		draw,
		-stealth,
		font=\footnotesize,
	},
}

\usepackage{pgfplots}
\pgfplotsset{compat=1.14}
\usepgfplotslibrary{fillbetween}

\usetikzlibrary{calc}
\usetikzlibrary{patterns} 

\usepackage[locale = US]{siunitx}		
\DeclareSIUnit\bar{bar}
\newcommand{\degreeKW}{\ensuremath{^\circ\text{CA}}}
\newcommand{\degreeKWnZOT}{\ensuremath{^\circ\text{CAaTDC}}}

\DeclareSIUnit{\degreeCrankAngle}{\degreeKW} 
\DeclareSIUnit{\degreeCrankAngleaTDC}{\degreeKWnZOT}
\DeclareSIUnit{\barperdegree}{\bar\per\degreeKW}

\usepackage{glossaries}

\usetikzlibrary{spy}
\usetikzlibrary {decorations.fractals,spy}

\usepackage{lmodern} 
\usepackage{amsmath,amssymb,amsfonts}
\usepackage[ruled,vlined]{algorithm2e}
\usepackage{graphicx}
\usepackage{textcomp}
\usepackage{xcolor}
\usepackage{booktabs} 
\usepackage{listings}
\usepackage{subcaption}
\usepackage{multirow}
\usepackage{balance}
\usepackage[super]{nth}
\usepackage[switch, mathlines]{lineno}
\usepackage{float}
\usepackage{url}
\usepackage[multiple]{footmisc}
\usepackage[export]{adjustbox}
\usepackage{xspace}
\usepackage{tcolorbox}
\usepackage{array}
\usepackage{gensymb}
\usepackage{tabularx}
\usepackage{longtable}
\usepackage{supertabular,booktabs}
\usepackage{hhline}
\usepackage{colortbl}
\usepackage{multirow}
\usepackage[short]{optidef}
\usepackage{wrapfig}

\usepackage{acronym}
\newcommand{\nox}{$\mathrm{NO_{x}}$}

\newcommand{\norm}[1]{\left\lVert#1\right\rVert}
\newcommand{\trans}{^\top}
\newcommand{\yimep}[1][]{y_{\text{IMEP}#1}}
\newcommand{\ynox}[1][]{y_{\text{NO}_x#1}}

\newcommand{\ymprr}[1][]{y_{\text{MPRR}#1}}
\newcommand{\uhydrogen}[1][]{u_{\text{DOI,hydrogen}#1}}
\newcommand{\udiesel}[1][]{u_{\text{DOI,diesel}#1}}

\newlength{\PlotWidth}
\newlength{\PlotHeight}	
\newlength{\DeltaHeight}

\newlength{\DeltaWidth}
\newlength{\Ypos}

\newlength{\DeltaHeightDashed}

\definecolor{mmpBlack}{rgb}{0, 0, 0}
\definecolor{mmpDarkBlue}{rgb}{0, 0.3294, 0.6235}
\definecolor{mmpMiddleBlue}{rgb}{0.251, 0.498, 0.7176}
\definecolor{mmpLightBlue}{rgb}{0.5569, 0.7294, 0.898}
\definecolor{mmpRed}{rgb}{0.8, 0.0275, 0.1176}
\definecolor{mmpGreen}{rgb}{0.3412, 0.6706, 0.1529}
\definecolor{mmpOrange}{rgb}{0.9647, 0.6588, 0}
\definecolor{mmpDarkGray}{rgb}{0.4, 0.4, 0.4} 
\definecolor{mmpMiddleDarkGray}{rgb}{0.6, 0.6, 0.6} 
\definecolor{mmpMiddleGray}{rgb}{0.75, 0.75, 0.75} 
\definecolor{mmpLightGray}{rgb}{0.9, 0.9, 0.9} 

\titleformat*{\section}{\normalsize\bfseries}
\titleformat*{\subsection}{\normalsize\itshape}

\begin{document}

\begin{flushright}
\begin{small}
\begin{tabular}{r@{}}
\hline
Proceedings of Combustion Institute -- Canadian Section\\
Spring Technical Meeting\\
University of Calgary\\
\underline{May 12-15, 2025}\\
\end{tabular}
\end{small}
\end{flushright}

\vspace{10pt}
\begin{center}
\begin{LARGE} Hybrid Reinforcement Learning and Model Predictive Control for Adaptive Control of Hydrogen-Diesel Dual-Fuel Combustion \end{LARGE}\\
\vspace{15pt}
\begin{large}J.~Bedei, A.~Winkler, J.~Andert
\end{large}
\vspace{5pt}\\
\normalsize
\textit{Teaching and Research Area Mechatronics in Mobile Propulsion, RWTH Aachen University, Aachen, Germany}
\vspace{5pt}\\
\begin{large} M.~McBain, C.R.~Koch and D.~Gordon\footnote{\noindent Corresponding author: dgordon@ualberta.ca}\end{large}
\vspace{5pt}\\
\normalsize
\textit{Department of Mechanical Engineering, University of Alberta, 116 St and 85 Ave, Edmonton, AB, Canada, T6G 2R3}

\end{center}

\begin{abstract}

Reinforcement Learning (RL) and Machine Learning Integrated - Model Predictive Control (ML-MPC) are promising approaches for optimizing hydrogen-diesel dual-fuel (H2DF) engine control, as they can effectively manage multiple-input multiple-output systems and control nonlinear processes. ML-MPC is advantageous for providing safe and optimal control inputs, ensuring the engine operates within predefined safety limits. In contrast, RL is distinguished by its adaptability to changing conditions through its learning-based approach. However, the practical implementation of either method alone poses significant challenges. RL requires high variance in control inputs during early learning phases, which can pose risks to the system by potentially executing unsafe actions, leading to mechanical damage. Conversely, ML-MPC relies on an accurate system model to generate optimal control inputs and has limited adaptability to system drifts, such as injector aging, which naturally occur in engine applications.

To address these limitations, this study proposes a hybrid RL and ML-MPC approach that uses a well-established ML-MPC framework while incorporating an RL agent to dynamically adjust the ML-MPC load tracking reference in response to changes in the environment. At the same time, the ML-MPC ensures that actions stay safe throughout the RL agent's exploration. To evaluate the effectiveness of this approach, fuel pressure is deliberately varied to introduce a model-plant mismatch between the ML-MPC and the engine test bench. The result of this mismatch is a root mean square error (RMSE) in indicated mean effective pressure (IMEP) of 0.57 bar when running the ML-MPC standalone. The experimental results demonstrate that RL successfully adapts to changing boundary conditions by altering the tracking reference while ML-MPC ensures safe control inputs, keeping pressure rise rates constrained. The quantitative improvement in load tracking by implementing RL is an RSME in IMEP of 0.44 bar. These findings highlight the feasibility of combining RL and ML-MPC for engine control and encourage further research on learning-based and adaptive control strategies in safety-critical real-world environments, contributing to the advancement of sustainable transportation technologies.
\end{abstract}

\section{Introduction}
In 2018, 12\% of all vehicles sold annually in Alberta were Class 8 heavy-duty vehicles. This small portion of vehicles contributed to 54\% of the fuel burned in the province. Therefore, focusing efforts on developing solutions that reduce emissions by using alternative fuels can impact the reduction of carbon emissions. Hydrogen can be used in diesel engines by incorporating dual-fuel combustion processes, referred to as hydrogen–diesel dual-fuel (H2DF). By substituting a portion of the diesel fuel with port-injected hydrogen, these engines can substantially reduce greenhouse gas emissions while utilizing existing production diesel engines. However, when burned in a combustion engine, hydrogen exhibits nonlinear and complex combustion characteristics. The potential for abnormal combustion, particularly at higher loads, can limit the amount of hydrogen that can be safely added. These factors introduce new challenges for engine control, necessitating advanced control strategies \cite{Anstrom} to ensure efficiency, safety, and accurate load tracking. 
Numerous control strategies have been proposed in the literature. One of these methods is Machine Learning Integrated - Model Predictive Control (ML-MPC), which has demonstrated encouraging results in Compression Ignition (CI) engines~\cite{Gordon.2022,Norouzi2022} and Homogeneous Charge Compression Ignition (HCCI) engines~\cite{Gordon2023PhD,Gordon2024ACC}. The four key advantages to using ML-MPC for H2DF are ~\cite{Gordon.2022}: 
\begin{enumerate*}[label={\arabic*)}]
    \item constraints for state, input, and output variables.
    \item uses a future horizon to improve optimal output for the set control law.
    \item capable of managing uncertainties in the system's delays, parameters, and nonlinearity.
    \item requires no complex physical plant model compared to traditional MPC.
\end{enumerate*}

However, ML-MPC faces a key challenge experienced by most advanced control methods: all potential system states cannot be considered. This can prevent the control system from reaching the desired output, as there could be key elements of the system causing disturbances that the controller cannot account for, resulting in a model-plant mismatch. An example of this is the aging of the system. A promising approach to address this challenge is the integration of Reinforcement Learning (RL) with ML-MPC, where RL enables the ML-MPC to maintain robust control performance despite such discrepancies. The combination of RL and ML-MPC has already shown great potential, for instance, in nonlinear quadcopter trajectory tracking~\cite{ghezzi2025} or building energy management \cite{Arroyo2022}.

RL is an ML algorithm that has achieved widespread success for engine control in various simulation environments \cite{Norouzi2023, Sharma2025} by learning optimal actions through exploration of system states and maximizing the long-term reward. However, the application of RL in real-world environments remains limited due to safety concerns, particularly the risks associated with high-variance exploratory control inputs in early learning phases. In \cite{Maldonado.2024}, RL was used to optimize fuel injection in spark-ignition engines, reducing cycle-to-cycle variability and improving fuel efficiency. In \cite{Bedei2025}, RL-based HCCI control with real-time safety monitoring to ensure safe operation is presented. However, standalone RL in real-world applications remains challenging due to limited stability guarantees and high data demands in safety-critical systems. Integrating RL with ML-MPC enables adaptation to system drift, while using ML-MPC’s stability and existing process model to ensure safe exploration.

In this work, the integration of RL and ML-MPC is presented in detail. The main goal is to adapt the ML-MPC to a changing plant, represented by changing the injection pressure to simulate injector drift (injector aging) and evaluate the controller's performance. The potential of RL used to enhance the performance of the ML-MPC for an internal combustion engine is experimentally shown. 
\section{Methodology}
\subsection{Experimental setup}


\begin{wrapfigure}{r}{0.5\textwidth}
\vspace{-1.8cm}
    \centering
    \includegraphics[width=0.48 \textwidth]{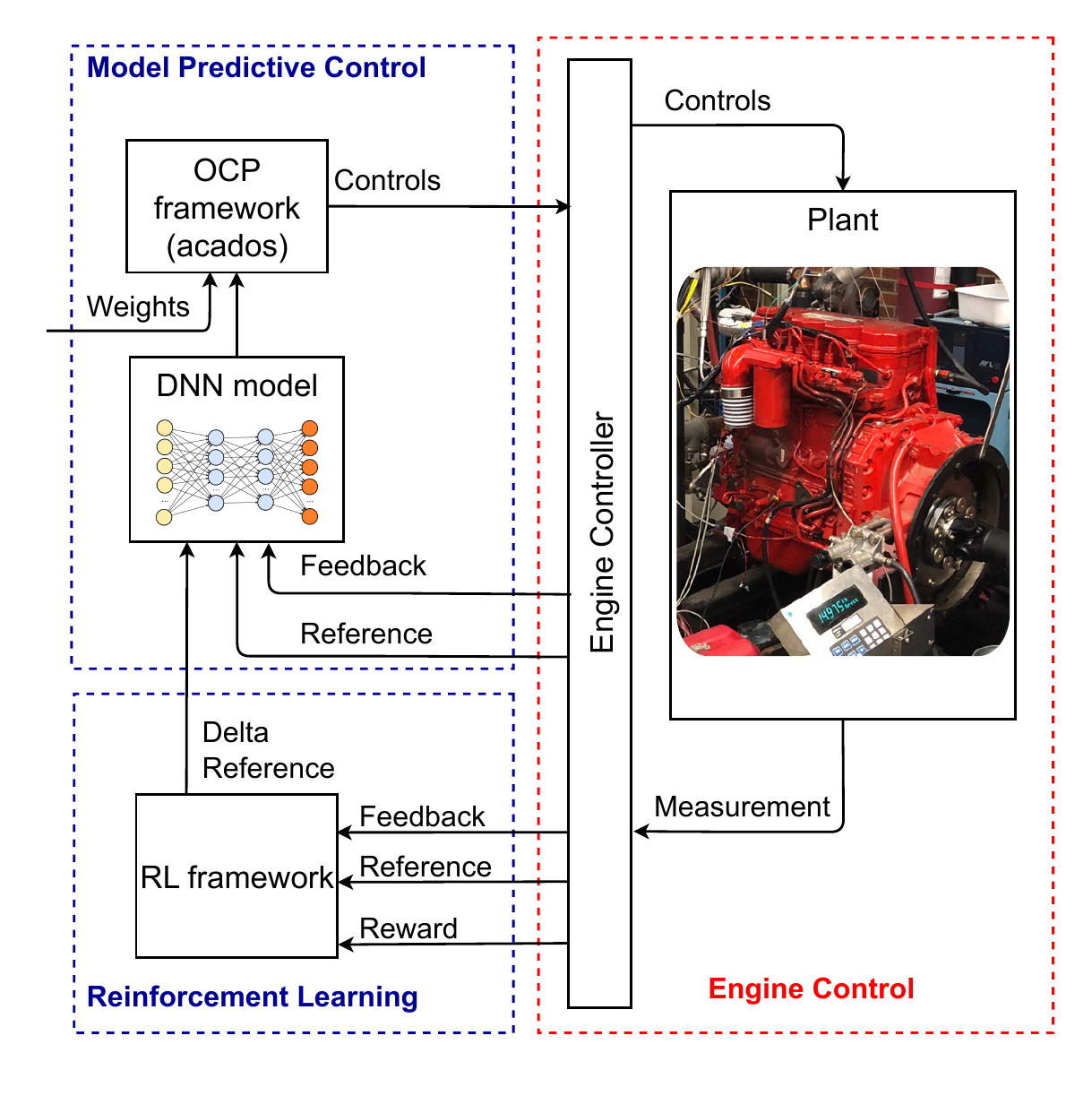}
    \caption{Hybrid Reinforcement Learning and Model Predictive Control Approach.}
    \label{fig:RLMPCStructure}
    \vspace{-0.8cm}
\end{wrapfigure}

The hybrid RL and ML-MPC approach, schematically shown in Figure~\ref{fig:RLMPCStructure}, is experimentally tested on a single cylinder of a Cummins QSB 4.5L 4-cylinder engine. A dSPACE MicroAutoBox II (MABX) rapid control prototyping (RCP) engine control unit (ECU) is used to control the engine and determine critical combustion metrics in real-time \cite{Gordon2024ACC}. The combustion metrics considered in this study consist of indicated mean effective pressure (IMEP), maximum pressure rise rate (MPRR) and nitrogen oxide emissions ($\text{NO}_{\text{x}}$) while the controls set by RL-MPC are the start of injection of diesel (SOI diesel), duration of injection of hydrogen gas (DOI hydrogen), and duration of injection of diesel (DOI Diesel). The combustion metrics are transmitted to a Raspberry Pi 400 (RPI), where the ML-MPC calculation is executed on an ARM Cortex A72 processor. In addition, the RPI is provided with a reference load profile. The reference load profile consists solely of a target IMEP. 

An RL agent is established on a standalone PC with a Linux operating system. The RL agent receives the same information as the RPI, along with a reward calculated on the MABX for each cycle. The agent calculates a reference load profile change, which is sent to the RPI to modify the ML-MPC reference trajectory, thus affecting the ML-MPC optimization and the resulting ECU control inputs for the next cycle.

\subsection{Reinforcement Learning}
\label{cha:RL}

RL is a machine learning approach in which an agent learns to make decisions through interaction with an environment - in this case, comprising both the ML-MPC and the real engine (Figure~\ref{fig:RLMPCStructure}). By receiving rewards~$r$ from the environment, the agent gradually improves its performance by training an actor, which defines the agent’s policy $\mu$ that aims to maximize the discounted cumulative reward $\sum^{\infty}_{i=0} \gamma^i \cdot r_i$. This so-called return is estimated by a critic, which is trained on the data generated during the agent’s exploration. In this study, both the actor and the critic are approximated by neural networks. To enable the application of RL in real time, the Open Source Learning and Experiencing Cycle Interface (LExCI) \cite{Badalian2024} is used, which facilitates the deployment of RL algorithms on embedded hardware.

The Twin Delayed Deep Deterministic Policy Gradient (TD3) algorithm is used in this study. Details on its internal mechanisms can be found in \cite{Fujimoto.2018}. The presented approach demonstrates how the flexibility and adaptability of RL can be harnessed to compensate for model–plant mismatches, thereby maintaining high ML-MPC performance even in the presence of system drifts. To support the ML-MPC algorithm, the action~$a$ of the RL agent provides an offset on the load reference $\Delta y_{\text{ref}, \text{IMEP},i}$ through its policy $\mu$ based on the state~$s_{i-1}$ that includes the combustion metrics as well as the desired IMEP of the current and last cycle.
\begin{equation}
\Delta y_{\text{ref}, \text{IMEP},i}=a_i=\mu\left(s_{i-1}\right), \quad 
s_{i-1} = \begin{bmatrix} y_{\text{IMEP},i-1}
\\ y_{\text{NO}_\text{x},i-1} \\ y_{\text{MPRR},i-1}\\y_{\text{ref}, \text{IMEP},i}\\y_{\text{ref}, \text{IMEP},i-1} \end{bmatrix} \in\mathbb{R}^5\end{equation}
The primary objective of the RL agent is to enhance the load tracking performance under system drift, while the ML-MPC ensures safety and addresses additional control objectives such as emission reduction and fuel efficiency. Consequently, the reward function focuses exclusively on the load tracking error. The reward for each combustion cycle $i$ is defined as:
\begin{equation}
r_{i}=  - 0.05\cdot\left(y_{\text{ref}, \text{IMEP},i}-y_{\text{IMEP},i}\right)^2 - 0.45\cdot\tanh\left(\left(y_{\text{ref}, \text{IMEP},i}-y_{\text{IMEP},i}\right)^2\right)
\end{equation}
The tanh term provides strong gradients near the target and saturates for large deviations, preventing exploding gradients and contributing to the numerical stability of the training using the gradient descent algorithm. Meanwhile, the small quadratic term ensures a non-zero gradient far from the target, maintaining motivation for the agent to adjust its policy~\cite{Bedei2025}.

\subsection{Machine learning integrated model predictive control}
To model the H2DF engine performance and emissions, a deep neural network~(DNN) with seven hidden layers (six Fully Connected (FC) layers and one GRU layer) similar to the model shown in \cite{Gordon.2022}. This model is then formulated using a nonlinear state-space representation to allow for integration into the MPC framework provided by \texttt{acados} as described in~\cite{Norouzi2022}. Thus, the discrete Optimal Control Problem (OCP) is defined based on the standard \texttt{acados} formulation \cite{Verschueren2021} as follows:

\begin{mini}
	{\substack{\Delta{u}_0, \dots, \Delta{u}_{N-1} \\ x_0, \dots, x_N \\ y_0, \dots, y_N}}%
	{\sum_{i=0}^{N} \norm{y_{ref,i}-y_i}^2_Q + \norm{\Delta{u}_i}^2_R}
	{}{}
	\addConstraint{x_0}{=\begin{bmatrix}x_{i},~u_{i-1}\end{bmatrix}\trans}
	\addConstraint{x_{i+1}}{= f(x_i, \Delta{u}_i)}{\quad\forall i\in\mathbb{H}\setminus N}
	\addConstraint{y_i}{= g(x_i, \Delta{u}_i)}{\quad\forall i\in\mathbb{H}}
	\addConstraint{u_\text{min}}{\le F_u \cdot u_i \le u_\text{max}}{\quad\forall i\in\mathbb{H}}
	\addConstraint{y_\text{min}}{\le F_y \cdot y_i \le y_\text{max}}{\quad\forall i\in\mathbb{H}}
	\label{eq:ocp}
\end{mini}
where $\mathbb{H}=\left\{0, 1, \dots, N\right\}$. Here $x_{i}$ are the internal model states (corresponding to the GRU: eight hidden states, $h_{i}$), ${y_{i}}$ the model outputs, and $u_{i}$ the model inputs. These are: 
\label{eq:ss_final}
\begin{align}
		x_{i} = \begin{bmatrix} h_{i-1} \end{bmatrix} \in\mathbb{R}^8, \quad 
		y_{i} = \begin{bmatrix} y_{\text{IMEP},i}  \\ y_{\text{NO}_{\text{x}},i} \\ y_{\text{MPRR},i} \end{bmatrix} \in\mathbb{R}^3, \quad 		
		u_{i} = \begin{bmatrix} y_{\text{IMEP},i-1} \\ u_{\text{SOI,diesel},i} \\ u_{\text{DOI,hydrogen},i} \\ u_{\text{DOI,diesel},i} \end{bmatrix}\in\mathbb{R}^4.
\end{align}

The reference $y_{ref}$ and the weighting matrix $Q$ are selected such that deviations from the requested IMEP are penalized while minimizing \nox~ emissions, duration of injected diesel fuel $\text{DOI,diesel}(i)$ and hydrogen $\text{DOI,hydrogen}(i)$ and change in control input $\Delta{u}$. Therefore, the specific cost function $J$ is specified as:

\begin{align}
	\label{eqn:cost_nmpc}
	J &= \sum_{i=0}^{N} \underbrace{\norm{y_{ref, \text{IMEP},i}+ \Delta y_{\text{ref}, \text{IMEP},i} - \yimep[,i]}^2_{q_\text{IMEP}}  }_{\text{Reference Tracking}}  +
    \underbrace{\norm{\ymprr[,i]}^2_{q_{\text{MPRR}}}}_{\text{Combustion Noise Reduction}}  + 
    \underbrace{\norm{\ynox[,i]}^2_{q_{\text{NO}_x}}}_{\text{Emission Reduction}}
    \\	
     &+    \underbrace{\norm{\udiesel[,i]}^2_{r_\text{DOI,fuel}} +\norm{\uhydrogen[,i]}^2_{r_\text{DOI,hydrogen}}}_{\text{Fuel / Hydrogen consumption reduction}} 
   +    \underbrace{\norm{\Delta{u}_i}^2_R}_{\text{Oscillation Reduction}}	 \nonumber
 \end{align}

One significant advantage of using ML-MPC for combustion control is the ability to impose constraints on inputs and outputs to ensure safe engine operation. $F_u$ and $F_y$ in Eq.~\ref{eq:ocp} are diagonal matrices that map the bounded outputs and inputs. The control outputs are limited to match the hardware used ($u_\text{min,max}$) while constraints imposed on the outputs ($y_\text{min,max}$) are used to guarantee safe engine operation. 

For the baseline ML-MPC testing, $\Delta y_{\text{ref}, \text{IMEP},i} = 0$, and when integrated with RL, the reference is modified by the RL agent as described in section~\ref{cha:RL}. 

\section{Results}

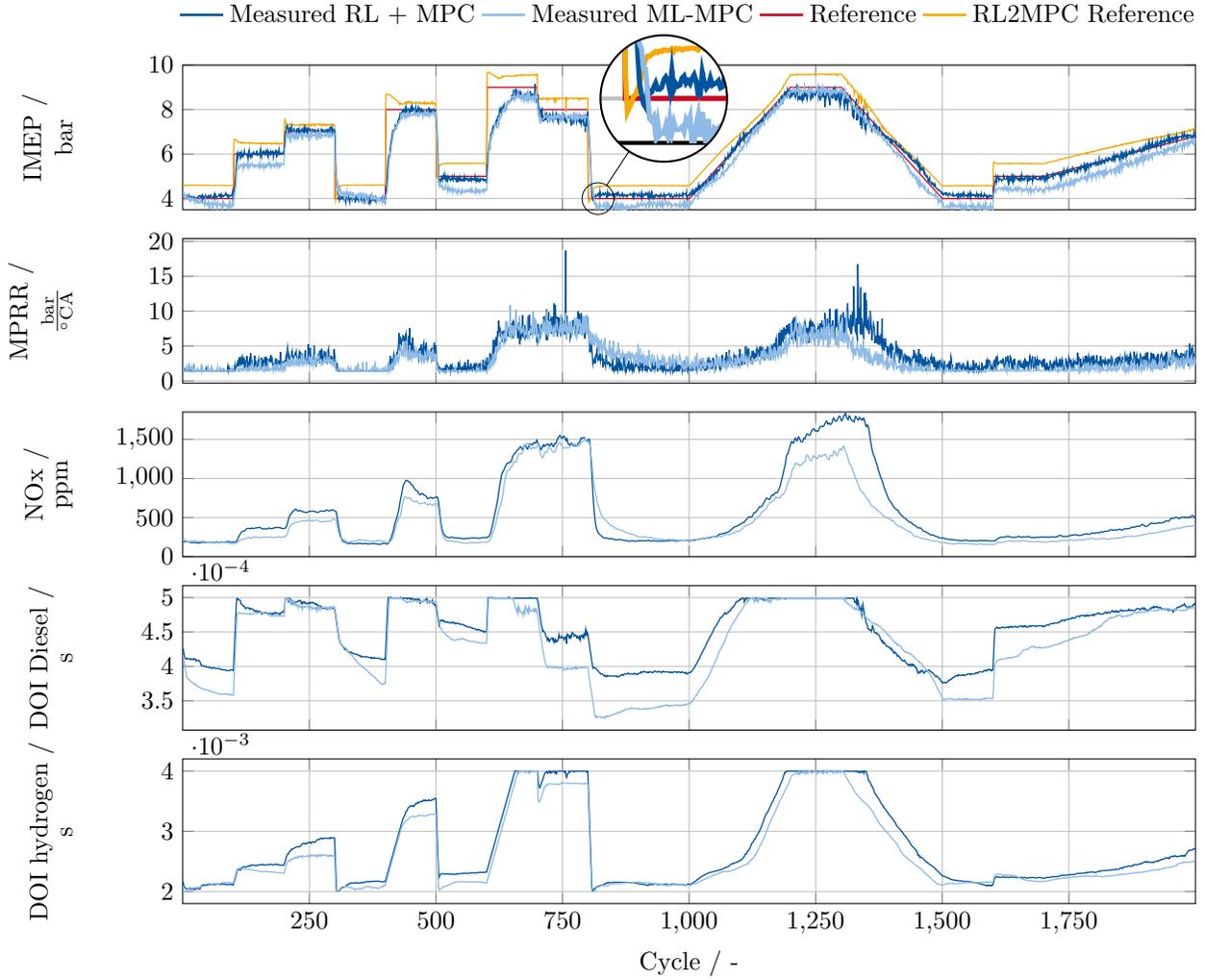
\begin{figure}[ht]
    \centering
    \begin{tikzpicture}[spy using outlines=
	{circle, magnification=4, connect spies}]
\setlength{\PlotWidth}{14cm} 
\setlength{\PlotHeight}{2cm} 
\setlength{\DeltaHeight}{0.4cm}
\setlength{\DeltaHeightDashed}{0.2cm}
\setlength{\DeltaWidth}{0.9cm}

\newcommand{\NumPlotY}{1}
\pgfmathsetlength{\Ypos}{-1*\NumPlotY*\PlotHeight - 1*\NumPlotY*\DeltaHeight}
\begin{axis}[%
	width=\PlotWidth,
	height=\PlotHeight,
	at={(0cm, \Ypos)},
	scale only axis,
	xmin=0,
	xmax=2000,
	xticklabel style={/pgf/number format/fixed},
	xtick={250, 500, 750, 1000, 1250, 1500, 1750},
	xticklabels = {},
	xlabel style={font=\color{white!15!black}},
	xlabel={},
	ymin=3.5,
	ymax=10,
	ylabel style={at={(-0.1,0.5)}},
    ylabel={\shortstack {IMEP / \\ \si{\bar}}},
    axis background/.style={fill=white},
	xmajorgrids,
	ymajorgrids,
	legend style={at={(0.5,1.2)},legend columns=4, legend cell align=center, align=center, anchor=south,  draw=none,fill=none},
	]
	\addplot[ color=mmpDarkBlue,line width=1pt] table {
		x y 
		0 1
		1 1
	};
	\addplot[ color=mmpLightBlue,line width=1pt] table {
		x y 
		0 1
		1 1
	};
	\addplot[ color=mmpRed,line width=1pt] table {
		x y 
		0 1
		1 1
	};
    \addplot[ color=mmpOrange,line width=1pt] table {
		x y 
		0 1
		1 1
	};
	
	\addplot[ color=mmpRed,line width=0.5pt,forget plot] table [x index=0, y index=1, col sep=comma] {Figures/Data/DataRLVSR10_episode21.csv};

   \addplot[ color=mmpOrange,line width=0.5pt,forget plot] table [x index=0, y index=12, col sep=comma] {Figures/Data/DataRLVSR10_episode21.csv};
   
	\addplot[ color=mmpDarkBlue,line width=0.5pt,forget plot] table [x index=0, y index=2, col sep=comma] {Figures/Data/DataRLVSR10_episode21.csv};


	\addplot[ color=mmpLightBlue,line width=0.5pt,forget plot] table [x index=0, y index=2, col sep=comma] {Figures/Data/MPCData_MCRL_0010MP0006.csv};

	\addlegendentry{Measured RL + MPC}
	\addlegendentry{Measured ML-MPC}
	\addlegendentry{Reference}
    \addlegendentry{RL2MPC Reference}

  \coordinate (spypoint) at (axis cs:820,4);
  \coordinate (magnifyglass) at (axis cs:950,8.5);
    
\end{axis}

  \spy [mmpBlack, size=1.75cm] on (spypoint)
   in node[fill=white] at (magnifyglass);

\renewcommand{\NumPlotY}{2}
\pgfmathsetlength{\Ypos}{-1*\NumPlotY*\PlotHeight - 1*\NumPlotY*\DeltaHeight}
\begin{axis}[%
	width=\PlotWidth,
	height=\PlotHeight,
	at={(0cm, \Ypos)},
	scale only axis,
	xmin=0,
	xmax=2000,
	xticklabel style={/pgf/number format/fixed},
	xtick={250, 500, 750, 1000, 1250, 1500, 1750},
	xticklabels = {},
	xlabel style={font=\color{white!15!black}},
	xlabel={},
	ylabel style={at={(-0.1,0.5)}},
	ylabel={\shortstack {MPRR / \\$\frac{\si{\bar}}{\si{\degreeKW}}$}},
	axis background/.style={fill=white},
	xmajorgrids,
	ymajorgrids,
	legend style={at={(0.02,2.6)},legend columns=3, legend cell align=left, align=left, anchor=north west, draw=none},
	]
	
    \addplot[ color=mmpDarkBlue,line width=0.5pt, forget plot] table [x index=0, y index=3, col sep=comma] {Figures/Data/DataRLVSR10_episode21.csv};


	\addplot[ color=mmpLightBlue,line width=0.5pt,forget plot] table [x index=0, y index=3, col sep=comma] {Figures/Data/MPCData_MCRL_0010MP0006.csv};

\end{axis}

\renewcommand{\NumPlotY}{3}
\pgfmathsetlength{\Ypos}{-1*\NumPlotY*\PlotHeight - 1*\NumPlotY*\DeltaHeight}
\begin{axis}[%
	width=\PlotWidth,
	height=\PlotHeight,
	at={(0cm, \Ypos)},
	scale only axis,
	xmin=0,
	xmax=2000,
	ymin=0,
	ymax=1850,
	xticklabel style={/pgf/number format/fixed},
	xtick={250, 500, 750, 1000, 1250, 1500, 1750},
	xticklabels = {},
	xlabel style={font=\color{white!15!black}},
	xlabel={},
	ylabel style={at={(-0.1,0.5)}},
	ylabel={\shortstack { NOx /  \\ ppm}},
	axis background/.style={fill=white},
	xmajorgrids,
	ymajorgrids,
	legend style={at={(0.02,2.6)},legend columns=2, legend cell align=left, align=left, anchor=north west, draw=none},
	]
	 	
    \addplot[ color=mmpDarkBlue,line width=0.5pt, forget plot] table [x index=0, y index=14, col sep=comma] {Figures/Data/DataRLVSR10_episode21.csv};
	\addplot[ color=mmpLightBlue,line width=0.5pt,forget plot] table [x index=0, y index=13, col sep=comma] {Figures/Data/MPCData_MCRL_0010MP0006.csv};
	
\end{axis}

\renewcommand{\NumPlotY}{4}
\pgfmathsetlength{\Ypos}{-1*\NumPlotY*\PlotHeight - 1*\NumPlotY*\DeltaHeight}
\begin{axis}[%
	width=\PlotWidth,
	height=\PlotHeight,
	at={(0cm, \Ypos)},
	scale only axis,
	xmin=0,
	xmax=2000,
	xticklabel style={/pgf/number format/fixed},
	xtick={250, 500, 750, 1000, 1250, 1500, 1750},
	xticklabels = {},
	xlabel style={font=\color{white!15!black}},
	xlabel={},
	ylabel style={at={(-0.1,0.5)}},
	ylabel={\shortstack  {DOI Diesel / \\ \si{\second}}},
	axis background/.style={fill=white},
	xmajorgrids,
	ymajorgrids,
	]
	\addplot[ color=mmpDarkBlue,line width=0.5pt,forget plot] table [x index=0, y index=4, col sep=comma] {Figures/Data/DataRLVSR10_episode21.csv};

    \addplot[ color=mmpLightBlue,line width=0.5pt,forget plot] table [x index=0, y index=4, col sep=comma] {Figures/Data/MPCData_MCRL_0010MP0006.csv};
	
\end{axis}

\renewcommand{\NumPlotY}{5}
\pgfmathsetlength{\Ypos}{-1*\NumPlotY*\PlotHeight - 1*\NumPlotY*\DeltaHeight}
\begin{axis}[%
	width=\PlotWidth,
	height=\PlotHeight,
	at={(0cm, \Ypos)},
	scale only axis,
	xmin=0,
	xmax=2000,
	xlabel style={font=\color{white!15!black}},
	xlabel={Cycle / -},
	xtick={250, 500, 750, 1000, 1250, 1500, 1750},
	ylabel style={at={(-0.1,0.5)}},
	ylabel={\shortstack  {DOI hydrogen / \\ \si{\second}}},
	axis background/.style={fill=white},
	xmajorgrids,
	ymajorgrids,
	]
	
	\addplot[ color=mmpDarkBlue,line width=0.5pt,forget plot] table [x index=0, y index=5, col sep=comma]		 {Figures/Data/DataRLVSR10_episode21.csv};

    \addplot[ color=mmpLightBlue,line width=0.5pt,forget plot] table [x index=0, y index=5, col sep=comma] {Figures/Data/MPCData_MCRL_0010MP0006.csv};
    
\end{axis}

\end{tikzpicture}
    \caption{Comparison of Control Results of MPC and Hybrid RL/ML-MPC approach with model-plant mismatch.}
    \label{fig:PlotResults}
\end{figure}

Initially, the ML-MPC's DNN is trained on data generated using a pseudorandom binary sequence on the controls while operating with a diesel fuel rail pressure of 1000 bar \cite{Gordon.2022}. To evaluate the hybrid RL/ML-MPC approach, the fuel rail pressure is reduced to 800 bar, introducing a model-plant mismatch. Figure~\ref{fig:PlotResults} illustrates the performance of the ML-MPC under this mismatch and the improvements achieved through the hybrid RL/ML-MPC approach.

A reference load profile is used to compare the controllers' performance. This reference load profile ranges from 4.5 bar to 9 bar IMEP with several step changes and linear changes in load. The reference load profile is illustrated in red in Figure \ref{fig:PlotResults}.

The actual IMEP of the ML-MPC with model-plant mismatch is shown in light blue. The deviations from the IMEP reference demonstrate that the ML-MPC struggles to track the profile and consistently underestimates the required fuel energy. This results from the rail pressure not being provided as an input to the ML-MPC, leaving the controller unaware of the resulting change in the environment.  The key performance indicator (KPI) used to evaluate load tracking performance is the root mean square error (RMSE) between the actual IMEP and the reference IMEP. Under this model-plant mismatch, the ML-MPC achieves an RMSE of 0.57 bar.

With a KPI baseline determined, the RL agent is deployed to reduce the RMSE in engine load. The RL agent is trained for a total of 56 episodes with 2000 combustion cycles each until it converges. Starting from randomly initialized actor and critic networks, the RL agent performs repeated runs of four training episodes with noise added to the policy output, followed by one validation episode without added noise. Each training episode uses random step changes for the IMEP reference between 4.5 and 9 bar, and the agent offsets its policy $\mu$ by Gaussian noise to explore the state-action space. The resulting data is used for network training to improve the agent's policy to maximize the cumulative reward.

After completing the 56 episodes, the hybrid RL/ML-MPC approach improved load tracking compared to the ML-MPC, achieving an RMSE of 0.44 bar. The adapted reference profile provided by the RL agent and the resulting IMEP trace are expressed in yellow and dark blue in Figure \ref{fig:PlotResults}. The agent introduces a consistent offset compared to the IMEP reference. This offset correlates directly with the underestimated IMEP in the initial model-plant mismatch in most instances. Two instances, from cycle 657 to 800 and from cycle 1190 to 1350, occur where the agent increases its reference profile offset in an attempt to improve load tracking. However, the load tracking performance cannot improve because both DOIs are constrained by the ML-MPC for safety reasons.

Moreover, the RL learned an overshooting feature at each step change in the load profile. An example of this overshoot feature can be observed in the zoomed-in bubble in Figure \ref{fig:PlotResults}. This has the potential to improve the load tracking performance; however, due to the DOIs rate constraints set by the ML-MPC for safety, the performance improvement is limited. Future research will exploit this potential to improve load tracking. 






\section{Conclusion}
In this study, RL is integrated with ML-MPC to control a H2DF combustion process within a single cylinder of a Cummins QSB 4.5L 4-cylinder engine. A model-plant mismatch is introduced by reducing the fuel rail pressure to simulate injector wear and aging. The ML-MPC without RL underestimates the required DOIs to meet the reference load profile. The RMSE between the actual IMEP of the ML-MPC and the target IMEP is 0.57 bar. An RL agent is then introduced to adjust the reference load profile to improve load tracking. The agent successfully develops an adapted reference with an offset from the initial reference corresponding to the absolute difference between the original model-mismatch ML-MPC IMEP and the target IMEP. Additionally, an overshoot feature is added to the reference, causing the ML-MPC to reach the target IMEP in fewer cycles. The resulting load tracking error with RL-MPC improves to an RMSE of 0.44 bar. Therefore, when introducing a plant-model mismatch, the RL and ML-MPC hybrid approach improves load tracking. However, there are three key limitations to implementing RL with ML-MPC: 
\begin{enumerate*}[label={\arabic*)}]
    \item the RL/ML-MPC hybrid approach may fail if the model and plant are substantially different.
    \item potential improved solutions can be limited by constraints set by the ML-MPC. These constraints within the ML-MPC may no longer be accurate. 
    \item limited potential to alter controller behavior due to the agent only adapting the reference load profile.
\end{enumerate*}
Though these limitations are present in this study, the results provide a foundation for future research into this hybrid approach to engine control. Future research will expand the RL agent's ability to alter the ML-MPC behavior by adjusting the ML-MPC's weights to improve load tracking. 









\section*{Acknowledgments}
The authors would like to support the previous work by Alexander Winkler has formed the foundations for this research. Funding for this work has been provided by the RWTH Aachen – University of Alberta Junior Research Fellowship and NSERC Discovery Grant RGPIN-2024-04990.


\footnotesize
\bibliographystyle{elsarticle-num}
\bibliography{references}  
\end{document}